\documentclass[sigconf]{acmart}

\usepackage{hyperref}
\usepackage{url}
\usepackage{amsfonts}
\usepackage{microtype}
\usepackage[center]{subfigure}
\usepackage{wrapfig}
\usepackage{amsthm}
\usepackage{enumitem}
\usepackage{algorithm,algorithmic}

\newtheorem{theorem}{Theorem}[section]

\newtheorem*{theorem*}{Theorem}
\newtheorem*{lemma*}{Lemma}
\newtheorem*{remark*}{Remark}
\newtheorem{assumption}[theorem]{Assumption}

\newtheorem{corollary}[theorem]{Corollary}

\newtheorem*{proposition*}{Proposition}

\AtBeginDocument{%
  }

\copyrightyear{2023} 
\acmYear{2023} 
\setcopyright{acmlicensed}\acmConference[WWW '23]{Proceedings of the ACM Web Conference 2023}{May 1--5, 2023}{Austin, TX, USA}
\acmBooktitle{Proceedings of the ACM Web Conference 2023 (WWW '23), May 1--5, 2023, Austin, TX, USA}
\acmPrice{15.00}
\acmDOI{10.1145/3543507.3583547}
\acmISBN{978-1-4503-9416-1/23/04}

\begin{document}

\title{Unlearning Graph Classifiers with Limited Data Resources}

\author{Chao Pan}
\authornote{Both authors contributed equally to this research.}
\affiliation{%
  \institution{University of Illinois, Urbana-Champaign}
  \city{Urbana}
  \state{Illinois}
  \country{USA}
}
\email{chaopan2@illinois.edu}

\author{Eli Chien}
\authornotemark[1]
\affiliation{%
  \institution{University of Illinois, Urbana-Champaign}
  \city{Urbana}
  \state{Illinois}
  \country{USA}
}
\email{ichien3@illinois.edu}

\author{Olgica Milenkovic}
\affiliation{%
  \institution{University of Illinois, Urbana-Champaign}
  \city{Urbana}
  \state{Illinois}
  \country{USA}
}
\email{milenkov@illinois.edu}

\renewcommand{\shortauthors}{Pan et al.}

\begin{abstract}
  As the demand for user privacy grows, controlled data removal (machine unlearning) is becoming an important feature of machine learning models for data-sensitive Web applications such as social networks and recommender systems. Nevertheless, at this point it is still largely unknown how to perform efficient machine unlearning of graph neural networks (GNNs); this is especially the case when the number of training samples is small, in which case unlearning can seriously compromise the performance of the model. To address this issue, we initiate the study of unlearning the Graph Scattering Transform (GST), a mathematical framework that is efficient, provably stable under feature or graph topology perturbations, and offers graph classification performance comparable to that of GNNs. Our main contribution is the first known nonlinear approximate graph unlearning method based on GSTs. Our second contribution is a theoretical analysis of the computational complexity of the proposed unlearning mechanism, which is hard to replicate for deep neural networks. Our third contribution are extensive simulation results which show that, compared to complete retraining of GNNs after each removal request, the new GST-based approach offers, on average, a $10.38$x speed-up and leads to a $2.6$\% increase in test accuracy during unlearning of $90$ out of $100$ training graphs from the IMDB dataset ($10$\% training ratio). Our implementation is available online at \url{https://doi.org/10.5281/zenodo.7613150}.
\end{abstract}



\maketitle

\section{Introduction}
\label{sec:intro}
Graph classification is a learning task that frequently arises in real-world Web applications related to social network analysis~\citep{mohammadrezaei2018identifying,fan2019graph,li2019semi,huang2021knowledge}, recommendation system development~\citep{ying2018graph,wu2020graph}, medical studies~\citep{li2019graph,mao2019medgcn}, and drug design~\citep{duvenaud2015convolutional,gaudelet2021utilizing}. While the availability of large sets of user training data has contributed to the deployment of modern deep learning models -- such as Graph Neural Networks (GNNs) -- for solving graph classification problems, deep learners mostly fail to comply with new data privacy regulations. Among them is the right of users to remove their data from the dataset and eliminate their contribution to all models trained on it. Such unlearning regulations\footnote{Existing laws on user data privacy include the European Union General Data Protection Regulation (GDPR), the California Consumer Privacy Act (CCPA), the Canadian Consumer Privacy Protection Act (CPPA), Brazilian General Data Protection Law (LGPD), and many others.} ensure the \emph{Right to be Forgotten,} and have to be taken into consideration during the model training process. We consider for the first time the problem of removing nodes from possibly different training graphs used in graph classification models. Examples where such requests arise include users (nodes) withdrawing from some interest groups (subgraphs) in social networks, and users deleting their browsing histories for online shopping, which corresponds to the removal of an entire graph from the training set. Note that only removing user data from a dataset is insufficient to guarantee the desired level of privacy, since the  training data can be potentially reconstructed from the trained models~\citep{fredrikson2015model,veale2018algorithms}. Naively, one can always retrain the model from scratch with the remaining dataset to ensure complete privacy; however, this is either computationally expensive or infeasible due to the high frequency of data removal requests~\citep{ginart2019making}. Another possible approach is to use differential privacy based methods to accommodate data removal during training. However, most state-of-the-art DP-GNNs focus on node classification instead of graph classification tasks, and it remains an open problem to design efficient GNNs for graph classification problem while preserving node-level privacy. Recently, the data removal problem has also motivated a new line of research on \emph{machine unlearning}~\citep{cao2015towards}, which aims to efficiently eliminate the influence of certain data points on a model. While various kinds of unlearning algorithms exist, most of the existing literature, especially the one pertaining to deep neural networks, does not discuss model training and unlearning in the limited training data regime; still, this is a very important data regime for machine learning areas such as few-shot learning~\citep{satorras2018few,wang2020generalizing}. Unlearning in this case may degrade the model performance drastically, and as a result, efficient and accurate graph classification and unlearning with limited training data is still an open problem. Throughout this paper, we use ``data removal'' and ``unlearning'' interchangeably.

Concurrently, a mathematical approach known as the Graph Scattering Transform (GST) has been used as a nontrainable counterpart of GNNs that can be applied to any type of trainable classification. GST iteratively applies graph wavelets and nonlinear activations on the input graph signals which may be viewed as forward passes of GNNs without trainable parameters. It has been shown that GSTs can not only remain stable under small perturbations of graph features and topologies~\citep{gama2019stability}, but also compete with many types of GNNs on solving different graph classification problems~\citep{gama2018diffusion,gao2019geometric,ioannidis2020pruned,pan2021spatiotemporal}. Furthermore, since all wavelet coefficients in GSTs are constructed analytically, GST is computationally more efficient and requires less training data compared to GNNs when combined with the same trainable classifiers. As a result, GSTs are frequently used in practice~\cite{min2020scattering,bianchi2021graph,li2021skeleton,pan2021spatiotemporal}, especially in the limited training data regime. Furthermore, it also allows for deriving theoretical analysis on nonlinear models for graph embedding, and the related conclusions could potentially shed some light on the design of GNNs that are more suitable for data removals.

\textbf{Our contributions.} We introduce the first \emph{nonlinear} graph learning framework based on GSTs that accommodates an \emph{approximate} unlearning mechanism with provable performance guarantees. Here, ``approximate'' refers to the fact that unlearning is not exact (as it would be for completely retrained models) but more akin to the parametrized notion of differential privacy~\cite{dwork2011differential,guo2020certified,sekhari2021remember} (see Section~\ref{sec:prelim} for more details). With the adoption of GSTs, we show that our nonlinear framework enables provable data removal (similar results are currently only available for linear models~\citep{guo2020certified,chien2022certified,chien2023efficient}) and provides theoretical unlearning complexity guarantees. These two problems are hard to tackle for deep neural network models like GNNs~\cite{xu2018how}. Our experimental results reveal that when trained with only $10\%$ samples in the dataset, our GST-based approach offers, on average, a $10.38$x speed-up and leads to a $2.6\%$ increase in test accuracy during unlearning of $90$ out of $100$ training graphs from the IMDB dataset, compared to complete retraining of a standard GNN baseline, Graph Isomorphism Network (GIN)~\cite{xu2018how}, after each removal request. Moreover, we also show that nonlinearities consistently improve the model performance on all five tested datasets, with an average increase of $3.5\%$ in test accuracy, which emphasizes the importance of analysis on nonlinear graph learning models.

\begin{figure*}[tb]
    \centering
    \includegraphics[width=0.9\linewidth]{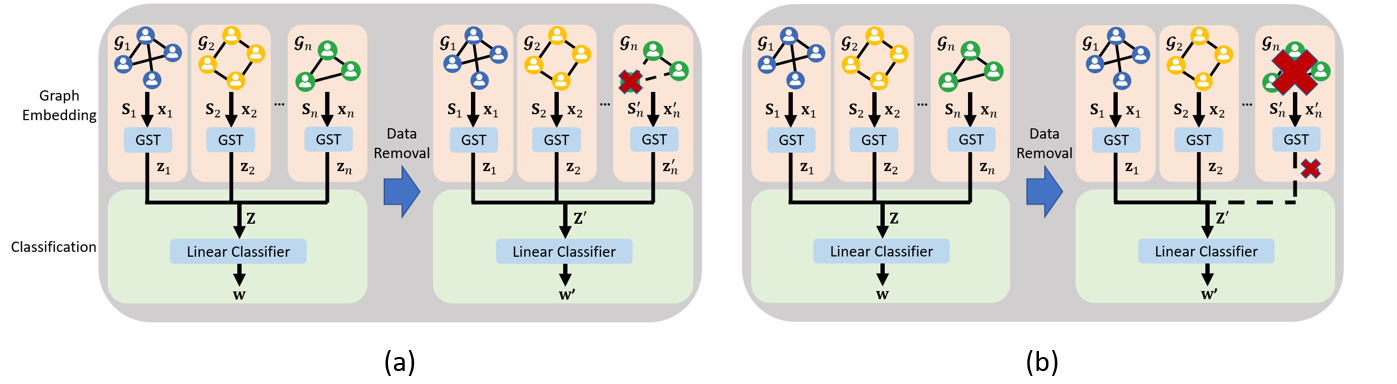}
    \vspace{-0.15in}
    \caption{The difference between removing one node from one training graph and completely removing one training graph, in the context of graph classification and when the \emph{graph embedding procedure} is nontrainable. Without loss of generality, we assume that the removal requests arise in the training graph $\mathcal{G}_n$. (a) The embedding of $\mathcal{G}_n$ changes from $\mathbf{z}_n$ to $\mathbf{z}_n^\prime$ and is used to train the new classifier $\mathbf{w}^\prime$. (b) The embedding of $\mathcal{G}_n$ is no longer used for training, and the underlying unlearning problem reduces to the one studied in~\cite{guo2020certified}. Since GST is a nontrainable feature extractor, the embeddings of all other graphs $\mathbf{z}_i,i\neq n$ remain the same for both (a) and (b).}
    \label{fig:unlearn_objective}
\end{figure*}

\vspace{-0.05in}
\section{Related Works}\label{sec:review}
\textbf{Machine unlearning.} The concept of machine unlearning was introduced in~\cite{cao2015towards}. Two unlearning criteria have been considered in the literature: \emph{Exact} and \emph{approximate unlearning}. Exact unlearning requires eliminating all information pertaining to the removed data, so that the unlearned model parameters have exactly the same ``distribution'' as the ones obtained by retraining from scratch. Examples of exact unlearning methods include sharding-based techniques~\citep{bourtoule2021machine} and quantization-based methods specialized for $k$-means clustering problems~\citep{ginart2019making,pan2023machine}. On the other hand, approximate unlearning only requires the distribution of the unlearned model parameters to be similar to retraining from scratch. One recent work~\cite{guo2020certified} introduced a probabilistic definition of unlearning motivated by differential privacy (DP)~\citep{dwork2011differential}, while~\cite{sekhari2021remember} performs a theoretical study of generalizations of unlearning and~\cite{golatkar2020eternal} addresses heuristic methods for deep learning models. Approximate unlearning offers a trade-off between model performance and privacy, as one is allowed to retain more relevant information compared to complete retraining so as not to cause serious performance degradation. This makes it especially desirable for limited training data regimes.

\textbf{Graph unlearning and DP-GNNs.} Only a handful of works have taken the initial steps towards machine unlearning of graphs. \cite{chen2021graph} proposes a sharding-based method for exact graph unlearning, while~\cite{chien2022certified,chien2023efficient} introduces approximate graph unlearning methods that come with theoretical (certified) guarantees. However, these works only focus on node classification tasks and are in general not directly applicable to graph classification. Moreover, the latter method~\cite{chien2023efficient} only considers linear model while the work described herein includes nonlinear graph transformations (i.e., GSTs). Note that although DP-GNNs~\citep{daigavane2021node,sajadmanesh2022gap} can also be used to achieve graph unlearning, they not only focus on node classification, but also require a high ``privacy cost'' to unlearn even \emph{one single node or edge} without causing significant performance degradation~\citep{chien2023efficient}. Similar observations regarding the relationship between DP and approximate unlearning were made in the context of unstructured unlearning~\cite{guo2020certified,sekhari2021remember}. Only one recent line of work considered differential privacy for graph classification~\cite{mueller2022differentially}, but the edit distance defined therein relates to the entire training graph instead of one or multiple nodes within the training graph. This approach hence significantly differs from our proposed data removal approach on graph classification tasks, and a more detailed comparative analysis is available in Section~\ref{sec:prelim}.

\textbf{Scattering transforms.} Convolutional neural networks (CNNs) use nonlinearities coupled with trained filter coefficients and are well-known to be hard to analyze theoretically~\citep{anthony2009neural}. As an alternative, \cite{mallat2012group, bruna2013invariant} introduced \emph{scattering transforms}, nontrainable versions of CNNs. Under certain conditions, scattering transforms offer excellent performance for image classification tasks and more importantly, they are analytically tractable. The idea of using specialized transforms has percolated to the graph neural network domain as well~\citep{gama2018diffusion, zou2020graph, gao2019geometric, pan2021spatiotemporal}. Specifically, the graph scattering transform (GST) proposed in~\cite{gama2018diffusion} iteratively performs graph filtering and applies element-wise nonlinear activation functions to input graph signals to obtain embeddings of graphs. It is computationally efficient compared to GNNs, and performs better than standard Graph Fourier Transform (GFT)~\cite{sandryhaila2013discrete} due to the adoption of nonlinearities.

\textbf{Few-shot learning.} Few-shot learning~\cite{wang2020generalizing} is a machine learning paradigm for tackling the problem of learning from a limited number of samples with supervised information. It was first proposed in the context of computer vision~\cite{fei2006one}, and has been successfully applied to many other areas including graph neural networks~\cite{satorras2018few} and social network analysis~\cite{li2020few}. At this point, it has not been addressed under the umbrella of unlearning, as in this case unlearning can drastically degrade the model performance. Consequently, efficient and accurate graph classification and unlearning within the few-shot learning setting remains an open problem.

\vspace{-0.05in}
\section{Preliminaries}\label{sec:prelim}
We reserve bold-font capital letters such as $\mathbf{X}$ for matrices, bold-font lowercase letters such as $\mathbf{x}$ for vectors, and calligraphic capital letters $\mathcal{G}_i=(V_i,E_i)$ for graphs, where the index $i\in[n]$ stands for the $i$-th training graph with vertex set $V_i$ and edge set $E_i$. Furthermore, $[\mathbf{x}]_i$ denotes the $i$-th element of vector $\mathbf{x}$. The norm $\|\cdot\|$ is by default the $\ell_2$ norm for vectors and the operator norm for matrices. We consider the graph classification problem as in~\cite{xu2018how}, where we have $n$ graphs with possibly different sizes for training (e.g., each graph $\mathcal{G}_i$ may represent a subgraph within the Facebook social networks). For a graph $\mathcal{G}_i$, the ``graph shift'' operator is denoted as $\mathbf{S}_i$. Graph shift operators include the graph adjacency matrix, graph Laplacian matrix, their normalized counterparts and others. For simplicity, this work focuses on $\mathbf{S}_i$ being the symmetric graph adjacency matrices. The graph node features are summarized in $\mathbf{X}_i\in\mathbb{R}^{g_i\times d_i}$, where $g_i=|V_i|$ denotes the number of nodes in $\mathcal{G}_i$ and $d_i$ the corresponding feature vector dimension. We also assume that $d_i=1$ for $\forall i\in[n]$ so that $\mathbf{X}_i$ reduces to a vector $\mathbf{x}_i$.

For each $\mathcal{G}_i$, we obtain an embedding vector $\mathbf{z}_i$ either via nontrainable, graph-based transforms such as PageRank~\citep{gleich2015pagerank}, GFT~\citep{shuman2013emerging} and GST~\citep{gama2019stability}, or geometric deep learning methods like GNNs~\citep{welling2016semi,hamilton2017inductive,xu2018how}. For GST, the length of $\mathbf{z}_i\in\mathbb{R}^d$ is determined by the number of nodes in the $L$-layer $J$-ary scattering tree; specifically, $d=\sum_{l=0}^{L-1} J^l, J,L\in\mathbb{N}^+$. All $\mathbf{z}_i$ are stacked vertically to form the data matrix $\mathbf{Z}$, which is used with the labels $\mathbf{y}$ to form the dataset $\mathcal{D}=(\mathbf{Z},\mathbf{y})$ for training a linear graph classifier. The loss equals 
\begin{align}\label{eq:loss}
L(\mathbf{w},\mathcal{D}) = \sum_{i=1}^n \left(\ell(\mathbf{w}^T\mathbf{z}_i,y_i) + \frac{\lambda }{2}\|\mathbf{w}\|^2\right),
\end{align}
where $\ell(\mathbf{w}^T\mathbf{z},y)$ is a convex loss function that is differentiable everywhere. We also write $\mathbf{w}^\star=\text{argmin}_\mathbf{w}L(\mathbf{w},\mathcal{D})$, and observe that the optimizer is unique whenever $\lambda>0$ due to strong convexity. For simplicity, we only consider the binary classification problem and point out that multiclass classification can be solved using well-known one-versus-all strategies.

\begin{figure}
    \centering
    \includegraphics[width=\linewidth]{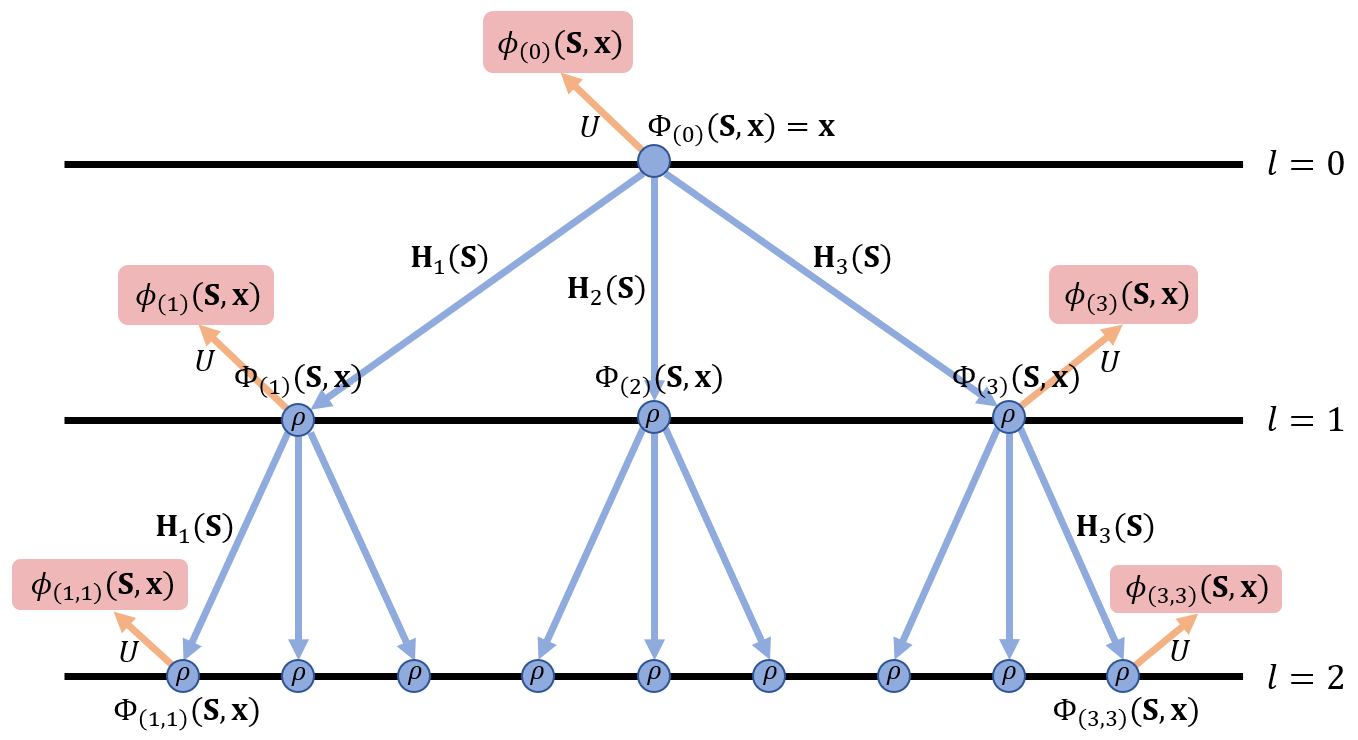}
    \vspace{-0.25in}
    \caption{The embedding procedure of GST with $L=J=3$. The scalar scattering coefficients $\phi_{p_j(l)}(\mathbf{S}, \mathbf{x})$ (red blocks) are concatenated to form the vector representation $\mathbf{z}$ of a graph.}
    \label{fig:gst}
\end{figure}

\textbf{Graph embedding via GST.} GST is a convolutional architecture including three basic units: 1) a bank of multiresolution wavelets $\{h_j\}_{j=1}^J$ (see Appendix~\ref{app:wavelet_formulation} for a list of common choices of graph wavelets); 2) a pointwise nonlinear activation function $\rho$ (i.e., the absolute value function); 3) A low-pass operator $U$ (i.e., averaging operator $\frac{1}{g_i}\mathbf{1}$). Note that the graph wavelets used in GST should form a frame~\citep{hammond2011wavelets,shuman2015spectrum} such that there exist $A\leq B$ and $\forall \mathbf{x}$, we have 
\begin{align}\label{eq:frame_property}
A^2\|\mathbf{x}\|^2 \leq \sum_{j=1}^J\left\|\mathbf{H}_j(\mathbf{S}) \mathbf{x}\right\|^2 \leq B^2\|\mathbf{x}\|^2,
\end{align}
where the graph shift operator $\mathbf{S}$ has the eigenvalue decomposition $\mathbf{S}=\mathbf{V}\mathbf{\Lambda}\mathbf{V}^T$ and $\mathbf{H}_j(\mathbf{S})=\mathbf{V}\mathbf{H}_j(\mathbf{\Lambda})\mathbf{V}^T=\mathbf{V}\text{diag}[h_j(\lambda_1),\ldots,h_j(\lambda_{n})]\mathbf{V}^T$. These elements are combined sequentially to generate a vector representation $\Phi(\mathbf{S}, \mathbf{x})$ of the input graph signal $\mathbf{x}$, as shown in Figure~\ref{fig:gst}. More specifically, the original signal $\mathbf{x}$ is positioned as the root node ($0$-th layer), and then $J$ wavelets are applied to each of the nodes from the previous layer, generating $J^l$ new nodes in the $l$-th layer to which the nonlinearity $\rho$ is applied. The scalar scattering coefficient $\phi_{p_j(l)}(\mathbf{S}, \mathbf{x})$ of node $p_j(l)$ is obtained by applying $U$ to the filtered signal $\Phi_{p_j(l)}(\mathbf{S}, \mathbf{x})=\mathbf{H}_{p_j(l)}(\mathbf{S})\mathbf{x}$, where the path $p_j(l)=(j_1,\ldots,j_l)$ denotes the corresponding node in the scattering tree. The coefficients are concatenated to form the overall representation of the graph $\Phi(\mathbf{S}, \mathbf{x})$, which is then used as the embedding of the graph $\mathbf{z}$. For a scattering transform with $L$ layers, the length of $\Phi(\mathbf{S}, \mathbf{x})$ is $d=\sum_{l=0}^{L-1} J^l$, which is independent of the number of nodes in the graph. For a more detailed description of GSTs, the reader is referred to~\cite{gama2019stability}.

\textbf{Removal requests.} The unlearning problem considered is depicted in Figure~\ref{fig:unlearn_objective}(a). Without loss of generality, we assume that the removal request arises for the $g_n$-th node in the training graph $\mathcal{G}_n$. We consider one removal request at a time, pertaining either to the removal of node features of one node ($\mathbf{x}_n\rightarrow \mathbf{x}_n^\prime$) or to the removal of the entire node ($\mathbf{S}_n\rightarrow \mathbf{S}_n^\prime, \mathbf{x}_n\rightarrow \mathbf{x}_n^\prime$) from one training graph (instead of the removal of one whole graph). In this case, the number of (training) graphs will remain the same ($n$), and we need to investigate the influence of the node removal on the graph embedding process (i.e., GST) to determine if removal significantly changes the embeddings of the affected graphs. Note that since GST is a nontrainable feature extractor, the embeddings of all unaffected training graphs will remain the same. On the other hand, if we are asked to unlearn the entire graph, the unlearning problem (see Figure~\ref{fig:unlearn_objective}(b)) reduces to the one studied in~\cite{guo2020certified}.

\textbf{Certified approximate removal.} Let $A$ be a (randomized) learning algorithm that trains on $\mathcal{D}$, the set of data points before removal, and outputs a model $h\in \mathcal{H}$, where $\mathcal{H}$ represents a chosen space of models. The data removal requests leads to a change from $\mathcal{D}$ to $\mathcal{D}^\prime$. Given a pair of parameters $(\epsilon,\delta)$, an unlearning algorithm $M$ applied to $A(\mathcal{D})$ is said to guarantee an $(\epsilon,\delta)$-certified approximate removal for $A$, where $\epsilon,\delta>0$ and $\mathcal{X}$ denotes the space of possible datasets, if $\forall \mathcal{T}\subseteq \mathcal{H}, \mathcal{D}\subseteq \mathcal{X}:$
\begin{align}\label{eq:CR_def}
    &\mathbb{P}\left(M(A(\mathcal{D}),\mathcal{D},\mathcal{D}^\prime)\in \mathcal{T}\right)
    \leq e^{\epsilon} \mathbb{P}\left(A(\mathcal{D}^\prime)\in \mathcal{T}\right)+\delta, \nonumber\\
    & \mathbb{P}\left(A(\mathcal{D}^\prime)\in \mathcal{T}\right)\leq e^{\epsilon}\mathbb{P}\left(M(A(\mathcal{D}),\mathcal{D},\mathcal{D}^\prime)\in \mathcal{T}\right)+\delta.
\end{align}
This definition is related to  $(\epsilon,\delta)$-DP~\citep{dwork2011differential} except that we are now allowed to update the model based on the updated dataset $\mathcal{D}^\prime$ under the certified approximate removal criterion. An $(\epsilon,\delta)$-certified approximate removal method guarantees that the updated model $M(A(\mathcal{D}),\mathcal{D},\mathcal{D}^\prime)$ is approximately the same from a probabilistic point of view as the model $A(\mathcal{D}^\prime)$ obtained by retraining from scratch on $\mathcal{D}^\prime$. Thus, any information about the removed data is approximately (but with provable guarantees) eliminated from the model. Note that exact removal corresponds to $(0,0)$-certified approximate removal. Ideally, we would like to design $M$ so that it satisfies Equation~(\ref{eq:CR_def}) with a predefined $(\epsilon,\delta)$ pair and has a complexity that is significantly smaller than that of complete retraining.

\textbf{Nonlinearities in the graph learning framework.} Previous work~\cite{chien2023efficient} on analyzing approximate unlearning of linearized GNNs focuses on node classification tasks within a single training graph. There, the training samples are node embeddings and the embeddings are correlated because of the graph convolution operation. The analysis of node classification proved to be challenging since propagation on graphs ``mixes'' node features, and thus the removal of even one feature/edge/node could lead to the change of embeddings for multiple nodes. The same technical difficulty exists for GNNs tackling graph classification tasks (graph embeddings are correlated), which limits the scope of theoretical studies of unlearning to GNNs that do not use nonlinear activations, such as SGC~\cite{wu2019simplifying}. Meanwhile, if we use nontrainable graph feature extractors (i.e., GSTs) for graph classification tasks to obtain the graph embeddings, the removal requests arising for one graph will not affect the embeddings of other graphs. This feature of GSTs not only makes the analysis of approximate unlearning tractable, but also allows one to introduce nonlinearities into the graph embedding procedure (i.e., the nonlinear activation function used in GSTs), which significantly improves the model performance in practice.

\section{Unlearning Graph Classifiers}\label{sec:gc_unlearn}
We now turn our attention to describing how the inherent stability properties of GSTs, which are nonlinear graph embedding methods, can aid in approximate unlearning without frequent retraining (a detailed discussion regarding how to replace GST with GNNs is available in Section~\ref{sec:discuss}).

Motivated by the unlearning approach described in~\cite{guo2020certified} for unstructured data, we design an unlearning mechanism $M$ that updates the trained model from $\mathbf{w}^\star$ to $\mathbf{w}^{\prime}$, the latter of which represents an approximation of the unique optimizer of $L(\mathbf{w},\mathcal{D}^\prime)$. Denote the Hessian of $L(\cdot,\mathcal{D}^\prime)$ at $\mathbf{w}^\star$ as $\mathbf{H}_{\mathbf{w}^\star} = \nabla^2 L(\mathbf{w}^\star, \mathcal{D}^\prime)$ and denote the gradient difference by $\Delta = \nabla L(\mathbf{w}^\star,\mathcal{D}) - \nabla L(\mathbf{w}^\star,\mathcal{D}^\prime)$. The update rule is $\mathbf{w}^{\prime} = \mathbf{w}^\star + \mathbf{H}_{\mathbf{w}^\star}^{-1}\Delta$, which can be intuitively understood as follows. Our goal is to achieve $\nabla L(\mathbf{w}^{\prime},\mathcal{D}^\prime)=0$ for the updated model. Using a Taylor series expansion we have
$$
\nabla L(\mathbf{w}^{\prime},\mathcal{D}^\prime)\approx \nabla L(\mathbf{w}^\star,\mathcal{D}^\prime) + \nabla^2 L(\mathbf{w}^\star,\mathcal{D}^\prime)(\mathbf{w}^{\prime}-\mathbf{w}^\star)=0.
$$
Therefore, we have
\begin{align}\label{eq:update_rule}
& \mathbf{w}^{\prime}-\mathbf{w}^\star=\left[\nabla^2 L(\mathbf{w}^\star,\mathcal{D}^\prime)\right]^{-1}\left[0-\nabla L(\mathbf{w}^\star,\mathcal{D}^\prime)\right] \notag \\
& \mathbf{w}^{\prime}=\mathbf{w}^\star + \left[\nabla^2 L(\mathbf{w}^\star,\mathcal{D}^\prime)\right]^{-1}\left[\nabla L(\mathbf{w}^\star,\mathcal{D})-\nabla L(\mathbf{w}^\star,\mathcal{D}^\prime)\right].
\end{align}
The last equality holds due to the fact that $\nabla L(\mathbf{w}^\star,\mathcal{D})=0$. When $\nabla L(\mathbf{w}^{\prime},\mathcal{D}^\prime)=0$, $\mathbf{w}^{\prime}$ is the unique optimizer of $L(\cdot, \mathcal{D}^\prime)$ due to strong convexity. If $\nabla L(\mathbf{w}^{\prime},\mathcal{D}^\prime) \neq 0$, some amount of information about the removed data point remains. One can show that the gradient residual norm $\|\nabla L(\mathbf{w}^{\prime},\mathcal{D}^\prime)\|$ determines the error of $\mathbf{w}^{\prime}$ when used to approximate the true minimizer of $L(\cdot, \mathcal{D}^\prime)$ again via Taylor series expansion.

We would like to point out that the update approach from~\cite{guo2020certified} originally designed for unstructured unlearning can be viewed as a special case of Equation~(\ref{eq:update_rule}) when the graph $\mathcal{G}_n$ needs to be unlearned completely. In this case, we have $\Delta=\nabla L(\mathbf{w}^\star,\mathcal{D})-\nabla L(\mathbf{w}^\star,\mathcal{D}^\prime)=\lambda \mathbf{w}^\star+\nabla\ell((\mathbf{w}^\star)^T\mathbf{z}_n,y_n)$, which is the same expression as the one used in~\cite{guo2020certified}. However, when we only need to unlearn part of the nodes in $\mathcal{G}_n$, $\Delta$ becomes $\Delta=\nabla\ell((\mathbf{w}^\star)^T\mathbf{z}_n,y_n)-\nabla\ell((\mathbf{w}^\star)^T\mathbf{z}^{\prime}_n,y_n)$, where $\mathbf{z}_n^{\prime}$ is obtained via GST computed on the remaining nodes in $\mathcal{G}_n$. The unlearning mechanism shown in Equation~(\ref{eq:update_rule}) can help us deal with different types of removal requests within a unified framework, and the main analytical contribution of our work is to establish bounds of the gradient residual norm for the generalized approach in the context of graph classification.

As discussed above, Equation~(\ref{eq:update_rule}) is designed to minimize the gradient residual norm $\|\nabla L(\mathbf{w}^{\prime},\mathcal{D}^\prime)\|$. However, the direction of the gradient residual $L(\mathbf{w}^{\prime},\mathcal{D}^\prime)$ may leak information about the training sample that was removed, which violates the goal of approximate unlearning. To address this issue,~\cite{guo2020certified} proposed to hide the real gradient residual by adding a linear noise term $\mathbf{b}^T\mathbf{w}$ to the training loss, a technique known as loss perturbation~\cite{chaudhuri2011differentially}. When taking the derivative of the noisy loss, the random noise $\mathbf{b}$ is supposed to ``mask'' the true value of the gradient residual so that one cannot infer information about the removed data. The corresponding approximate unlearning guarantees for the proposed unlearning mechanism can be established by leveraging Theorem~\ref{thm:app_thm3} below.

\begin{theorem}[Theorem 3 from~\cite{guo2020certified}]\label{thm:app_thm3} Denote the noisy training loss by $L_{\mathbf{b}}(\mathbf{w},\mathcal{D}) = \sum_{i=1}^n \left(\ell(\mathbf{w}^T\mathbf{z}_i,y_i) + \frac{\lambda }{2}\|\mathbf{w}\|^2\right)+\mathbf{b}^T\mathbf{w}$, and let $A$ be the learning algorithm that returns the unique optimum of $L_{\mathbf{b}}(\mathbf{w},\mathcal{D})$. 
Suppose that $\mathbf{w}^\prime$ is obtained by the unlearning procedure $M$ and that $\|\nabla L(\mathbf{w}^{\prime},\mathcal{D}^\prime)\|\leq \epsilon^\prime$ for some computable bound $\epsilon^\prime > 0$. If $\mathbf{b}\sim \mathcal{N}(0,c \epsilon^\prime/\epsilon)^d$ is normally distributed with some constant $\epsilon,c>0$, then $M$ satisfies Equation~(\ref{eq:CR_def}) with $(\epsilon,\delta)$ for algorithm $A$ applied to $\mathcal{D}^\prime$, where $\delta = 1.5\cdot e^{-c^2/2}$. 
\end{theorem}
Hence, if we can appropriately bound the gradient residual norm $\|\nabla L(\mathbf{w}^\prime,\mathcal{D}^\prime)\|$ for graph classification problems,
we can show that the unlearning mechanism ensures an $(\epsilon,\delta)$-certified approximate removal. For the analysis, we need the following assumptions on the loss function $\ell$. These assumptions naturally hold for commonly used linear classifiers such as linear regression and logistic regression (see Section~\ref{sec:discuss}).

\begin{assumption}\label{asp:guo}
There exist constants $C_1,C_2,\gamma_1,\gamma_2$ such that for $\forall i\in[n]$ and $\mathbf{w}\in\mathbb{R}^d$: 1) $\|\nabla\ell(\mathbf{w}^T\mathbf{z}_i,y_i)\|\leq C_1$; 2) $|\ell^\prime(\mathbf{w}^T\mathbf{z_i},y_i)|\leq C_2$; 3) $\ell^\prime$ is $\gamma_1$-Lipschitz; 4) $\ell^{\prime\prime}$ is $\gamma_2$-Lipschitz; 5) it is always possible to rescale $\mathbf{x}_i$ for graph $\mathcal{G}_i$ so that $|[\mathbf{x}_i]_j| \leq 1, \forall j\in[g_i]$.
\end{assumption}

We show next that the gradient residual norm for graph classification can be bounded for both types of removal requests. 

\begin{theorem}\label{thm:worst_case_single_feature}  
Suppose that Assumptions~\ref{asp:guo} hold, and that the difference between the original dataset $\mathcal{D}=(\mathbf{Z},\mathbf{y})$ and the updated dataset $\mathcal{D}^\prime=(\mathbf{Z}^\prime,\mathbf{y})$ is in the embedding of the $n$-th training graph, which equals $\mathbf{z}_n^\prime=\Phi(\mathbf{S}_n,\mathbf{x}_n^\prime)$ with $[\mathbf{x}_n^\prime]_{g_n}=0$. Let $B$ be the frame constant for the graph wavelets used in GST (see Equation~(\ref{eq:frame_property})). Then
\begin{equation}\label{eq:worst_case_single_feature}
    \|\nabla L(\mathbf{w}^\prime,\mathcal{D}^\prime)\|\leq \frac{\gamma_2 F^3}{\lambda^2 n}\min\left\{4C_1^2,\frac{(\gamma_1 C_1 F^2 + \lambda C_2 F)^2}{\lambda^2 g_n}\right\},
\end{equation}
where $F=\sqrt{\sum_{l=0}^{L-1} B^{2l}}$. For tight energy preserving wavelets we have $B=1,F=\sqrt{L}$.
\end{theorem}

The proof of Theorem~\ref{thm:worst_case_single_feature} can be found in Appendix~\ref{app:pf_single_feature_worst}. The key idea is to use the stability property of GSTs, as we can view the change of graph signal from $\mathbf{x}_n$ to $\mathbf{x}_n^\prime$ as a form of signal perturbation. The stability property ensures that the new embedding $\Phi(\mathbf{S}_n,\mathbf{x}_n^\prime)$ does not deviate significantly from the original one $\Phi(\mathbf{S}_n,\mathbf{x}_n)$, and allows us to establish the upper bound on the norm of gradient residual. Note that the second term on the RHS in Equation~(\ref{eq:worst_case_single_feature}) decreases as the size $g_n$ of the graph $\mathcal{G}_n$ increases. This is due to the averaging operator $U$ used in GSTs, and thus the graph embedding is expected to be more stable under signal perturbations for large rather than small graphs.

Next, we consider a more common scenario where an entire node in $\mathcal{G}_n$ needs to be unlearned. This type of request frequently arises in graph classification problems for social networks, where unlearning one node corresponds to one user withdrawing from one or multiple social groups. In this case, we have the following bound on the gradient residual norm.

\begin{theorem}\label{thm:worst_case_single}
Suppose that Assumptions~\ref{asp:guo} hold, and that both the features and all edges incident to the $g_n$-th node in $\mathcal{G}_n$ have to be unlearned. Then
\begin{equation}\label{eq:worst_case_single}
    \|\nabla L(\mathbf{w}^\prime,\mathcal{D}^\prime)\| \leq \frac{4\gamma_2 C_1^2 F^3}{\lambda^2 n}, F=\sqrt{\sum_{l=0}^{L-1} B^{2l}}.
\end{equation}
\end{theorem}
\begin{remark*}
In this case, the norm $\|\mathbf{z}_n^\prime-\mathbf{z}_n\|$ capturing the change in the graph embeddings obtained via GST is proportional to the norm of the entire graph signal $\|\mathbf{x}_n\|$. The second term within the $\min$ function is independent on $g_n$ and likely to be significantly larger than the first term. Thus, we omit the second term in Equation~(\ref{eq:worst_case_single}). More details are available in Appendix~\ref{app:pf_single_worst}.
\end{remark*}

\textbf{Batch removal.} The update rule in Equation~(\ref{eq:update_rule}) naturally supports removing multiple nodes from possibly 
different graphs at the same time. We assume that the number of removal requests $m$ at one time instance is smaller than the minimum size of a training graph, i.e., $m < \min_i g_i$, to exclude the trivial case of unlearning an entire graph. In this setting, we have the following upper bound on the gradient residual norm, as described in Corollary~\ref{coro:feature_worst_case_batch} and~\ref{coro:worst_case_batch}. The proofs are delegated to Appendix~\ref{app:pf_coro_worst}.

\begin{corollary}\label{coro:feature_worst_case_batch}
Suppose that Assumptions~\ref{asp:guo} hold, and that $m$ nodes from $n$ graphs have requested feature removal. Then
\begin{equation}\label{eq:feature_worst_case_batch}
    \|\nabla L(\mathbf{w}^\prime,\mathcal{D}^\prime)\|\leq \frac{\gamma_2 m^2 F^3}{\lambda^2 n}\min\left\{4C_1^2,\frac{(\gamma_1 C_1 F^2 + \lambda C_2 F)^2}{\lambda^2 g_n}\right\},
\end{equation}
where $F=\sqrt{\sum_{l=0}^{L-1} B^{2l}}$.
\end{corollary}

\begin{corollary}\label{coro:worst_case_batch}  
Suppose that Assumptions~\ref{asp:guo} hold, and that $m$ nodes from $n$ graphs have requested entire node removal. Then
\begin{equation}\label{eq:worst_case_batch}
    \|\nabla L(\mathbf{w}^\prime,\mathcal{D}^\prime)\| \leq \frac{4\gamma_2 m^2 C_1^2 F^3}{\lambda^2 n}, F=\sqrt{\sum_{l=0}^{L-1} B^{2l}}.
\end{equation}
\end{corollary}

\textbf{Data-dependent bounds.} The upper bounds in Theorems~\ref{thm:worst_case_single_feature} and~\ref{thm:worst_case_single} contain a constant factor $1/\lambda^2$ which may be large when $\lambda$ is small and $n$ is moderate. This issue arises due to the fact that those bounds correspond to the worst case setting for the gradient residual norm. Following an approach suggested in~\cite{guo2020certified}, we also investigated data-dependent gradient residual norm bounds which can be efficiently computed and are much tighter than the worst-case bound. Note that these are the bounds we use in the online unlearning procedure of Algorithm~\ref{alg:unlearning} for simulation purposes.
\begin{theorem}\label{thm:data_dependent}  
Suppose that Assumptions~\ref{asp:guo} hold. For both single and batch removal setting, and for both feature and node removal requests, one has
\begin{equation}\label{eq:data_dependent_single}
    \left\|\nabla L\left(\mathbf{w}^{\prime}, \mathcal{D}^{\prime}\right)\right\| \leq \gamma_2 F \left\|\mathbf{Z}^{\prime}\right\|\left\|\mathbf{H}_{\mathbf{w}^\star}^{-1} \Delta\right\|\left\|\mathbf{Z}^{\prime} \mathbf{H}_{\mathbf{w}^\star}^{-1} \Delta\right\|,
\end{equation}
where $\mathbf{Z}^\prime$ is the data matrix corresponding to the updated dataset $\mathcal{D}^\prime$.
\end{theorem}

\textbf{Algorithmic details.} The pseudo-codes for training unlearning models, as well as the single node removal procedure are described below. During training, a random linear term $\mathbf{b}^T \mathbf{w}$ is added to the training loss. The choice of standard deviation $\alpha$ determines the privacy budget $\alpha\epsilon/\sqrt{2\log(1.5/\delta)}$ that is used in Algorithm~\ref{alg:unlearning}. During unlearning, $\beta$ tracks the accumulated gradient residual norm. If it exceeds the budget, then $(\epsilon,\delta)$-certified approximate removal for $M$ is no longer guaranteed. In this case, we completely retrain the model using the updated dataset $\mathcal{D}^\prime$.

\begin{algorithm}
  \caption{Training Procedure}
  \label{alg:training}
  \begin{algorithmic}[1]
    \STATE \textbf{input:} Training dataset $\mathcal{D}=\{\mathbf{z}_i\in\mathbb{R}^{d},y_i\}_{i=1}^n$, loss $\ell$, parameters $\alpha,\lambda > 0$.
    \STATE Sample the noise vector $\mathbf{b}\sim\mathcal{N}(0,\alpha^2)^d$.
    \STATE $\mathbf{w}^\star = \arg\min_{\mathbf{w}\in\mathbb{R}^d}\sum_{i=1}^n \left(\ell(\mathbf{w}^T\mathbf{z}_i,y_i) + \frac{\lambda }{2}\|\mathbf{w}\|^2\right)+\mathbf{b}^T\mathbf{w}$.
    \RETURN $\mathbf{w}^\star$.
  \end{algorithmic}
\end{algorithm}

\vspace{-0.1in}
\begin{algorithm}
  \caption{Unlearning Procedure}
  \label{alg:unlearning}
  \begin{algorithmic}[1]
    \STATE \textbf{input:} Training graphs $\mathcal{G}_i$ with features $\mathbf{x}_i$, graph shift operator $\mathbf{S}_i$ and label $y_i$, loss $\ell$, removal requests $\mathcal{R}_m=\{r_1,r_2,\ldots\}$, parameters $\epsilon, \delta, \gamma_2, \alpha,\lambda,F > 0$.
    \STATE Compute the graph embeddings $\mathbf{z}_i$ via GST$(\mathbf{x}_i,\mathbf{S}_i)$. Initiate the training set $\mathcal{D}=\{\mathbf{z}_i\in\mathbb{R}^{d},y_i\}_{i=1}^n$.
    \STATE Compute $\mathbf{w}$ using Algorithm~\ref{alg:training} ($\mathcal{D}, \ell, \alpha, \lambda$).
    \STATE Set the accumulated gradient residual norm to $\beta=0$.
    \FOR{$r \in \mathcal{R}_m$}
      \STATE In $\mathbf{x}_r^\prime$, set the feature of a node to be removed to $0$. Update $\mathbf{S}_{r}^\prime$ if the entire node is to be removed. 
      \STATE Compute the new graph embedding $\mathbf{z}_r^\prime$ with GST$(\mathbf{x}_r^\prime,\mathbf{S}_r^\prime)$.
      \STATE Update the training set $\mathcal{D}^\prime$ and $\mathbf{Z}^\prime$.
      \STATE Compute $\Delta=\nabla L\left(\mathbf{w}, \mathcal{D}\right)-\nabla L\left(\mathbf{w}, \mathcal{D}^{\prime}\right), \mathbf{H}=\nabla^{2} L\left(\mathbf{w},\mathcal{D}^{\prime}\right)$.
      \STATE Update the accumulated gradient residual norm as $\beta = \beta + \gamma_2 F\|\mathbf{Z}^\prime\|\|\mathbf{H}^{-1}\Delta\|\|\mathbf{Z}^\prime\mathbf{H}^{-1}\Delta\|$.
      \IF{$\beta > \alpha\epsilon/\sqrt{2\log(1.5/\delta)}$}
      \STATE Recompute $\mathbf{w}$ using Algorithm~\ref{alg:training} ($\mathcal{D}^\prime, \ell, \alpha, \lambda$), $\beta=0$.
      \ELSE
      \STATE $\mathbf{w}=\mathbf{w}+\mathbf{H}^{-1}\Delta$.
      \ENDIF
      \STATE $\mathcal{D}=\mathcal{D}^\prime$.
    \ENDFOR
    \RETURN $\mathbf{w}$.
  \end{algorithmic}
\end{algorithm}

\vspace{-0.05in}
\section{Discussion}\label{sec:discuss}
\textbf{Commonly used loss functions.} For linear regression, the loss function is $\ell(\mathbf{w}^T\mathbf{z}_i,y_i)=(\mathbf{w}^T\mathbf{z}_i-y_i)^2$, while $\nabla^2\ell(\mathbf{w}^T\mathbf{z}_i,y_i)=\mathbf{z}_i\mathbf{z}_i^T$, which does not depend on $\mathbf{w}$. Therefore, it is possible to have $\|\nabla L(\mathbf{w}^\prime,\mathcal{D}^\prime)\|=0$ based on the proof in Appendix~\ref{app:pf_single_feature_worst}. This observation implies that our unlearning procedure $M$ is a $(0,0)$-certified approximate removal method when linear regression is used as the linear classifier module. Thus, the exact values of $C_1,C_2,\gamma_1,\gamma_2$ are irrelevant for the performance guarantees for $M$.

For binary logistic regression, the loss function is defined as $\ell(\mathbf{w}^T\mathbf{z}_i,y_i) = -\log(\sigma(y_i\mathbf{w}^T\mathbf{z}_i))$, where $\sigma(x) = 1/(1+\exp(-x))$ denotes the sigmoid function. As shown in~\cite{guo2020certified}, the assumptions (1) and (4) in~\ref{asp:guo} are satisfied with $C_1=1$ and $\gamma_2 = 1/4$. We only need to show that (2) and (3) of~\ref{asp:guo} hold as well. Observe that $\ell'(x, y) = \sigma(xy)-1$. Since the sigmoid function $\sigma(\cdot)$ is restricted to lie in $[0,1]$, $|\ell'|$ is bounded by $1$, which means that our loss satisfies (2) in~\ref{asp:guo} with $C_2=1$. Based on the Mean Value Theorem, one can show that $\sigma(x)$ is $\max_{x\in \mathbb{R}}|\sigma^\prime(x)|$-Lipschitz. With some simple algebra, one can also prove that $\sigma^\prime(x) = \sigma(x)(1-\sigma(x)) \Rightarrow \max_{x\in \mathbb{R}}|\sigma^\prime (x)| = 1/4$. Thus the loss satisfies assumption (3) in~\ref{asp:guo} as well, with $\gamma_1=1/4$. For multiclass logistic regression, one can adapt the one-versus-all strategy which leads to the same result.

Note that it is also possible to use other loss functions such as linear SVM in Equation~(\ref{eq:loss}) with regularization. Choosing appropriate loss function for different applications could be another interesting future direction of this work.

\textbf{Reducing the complexity of recomputing graph embeddings.} Assume that the removal request arises in graph $\mathcal{G}_n$. For the case of feature removal, since we do not need to update the graph shift operator $\mathbf{S}_n$, we can reuse the graph wavelets $\{\mathbf{H}_j(\mathbf{S}_n)\}_{j=1}^J$ computed before the removal to obtain the new embedding $\mathbf{z}_n^\prime$, which is with complexity $O(d g_n^2)$.

For the case of complete node removal, we do need to update the graph wavelets $\{\mathbf{H}_j(\mathbf{S}_n^\prime)\}_{j=1}^J$ based on the updated $\mathbf{S}_n^\prime$. In general, the complexity of computing $\mathbf{z}_n^\prime$ in this case equals $O(d g_n^3)$, as we need to compute the eigenvalue decomposition of $\mathbf{S}_n^{\prime}$ and perform matrix multiplications multiple times. This computational cost may be too high when the size $g_n$ of $\mathcal{G}_n$ is large. There are multiple methods to reduce this computational complexity, which we describe next.

If the wavelet kernel function $h(\lambda)$ is a polynomial function, we can avoid the computation of the eigenvalue decomposition of $\mathbf{S}_n^\prime$ by storing the values of all $\{\mathbf{S}_n^k\}_{k=1}^K$ in advance during initial training, where $K$ is the degree of $h(\lambda)$. For example, if $h(\lambda)=\lambda - \lambda^2$, we have $\mathbf{H}(\mathbf{S}_n^\prime)=\mathbf{V}^\prime(\mathbf{\Lambda}^\prime - {\mathbf{\Lambda}^\prime}^2){\mathbf{V}^\prime}^T=\mathbf{S}_n^\prime-{\mathbf{S}_n^\prime}^2$. Note that we can write the new graph shift operator as $\mathbf{S}_n^\prime=\mathbf{S}_n+\mathbf{E}\mathbf{S}_n+\mathbf{S}_n\mathbf{E}$, where $\mathbf{E}$ is a diagonal matrix (i.e., if we remove the $g_n$-th node in $\mathcal{G}_n$, we have $\mathbf{E}=\text{diag}[0,\ldots,0,-1]$). In this case, ${\mathbf{S}_n^\prime}^2$ can be found as
\vspace{-0.05in}
$$
{\mathbf{S}_n^\prime}^2=\mathbf{S}_n^2+2\mathbf{S}_n\mathbf{E}\mathbf{S}_n+\mathbf{S}_n^2\mathbf{E}+\mathbf{E}\mathbf{S}_n^2+\mathbf{E}\mathbf{S}_n\mathbf{E}\mathbf{S}_n+\mathbf{E}\mathbf{S}_n^2\mathbf{E}+\mathbf{S}_n\mathbf{E}^2\mathbf{S}_n+\mathbf{S}_n\mathbf{E}\mathbf{S}_n\mathbf{E}.
$$
Thus, if we can store the values of all $\{\mathbf{S}_n^k\}_{k=1}^K$ in advance during initial training, we can reduce the complexity of computing $\{{\mathbf{S}_n^\prime}^k\}_{k=1}^K$ to $O(g_n^2)$, due to the fact that whenever $\mathbf{E}$ is involved in a matrix multiplication (i.e., $\mathbf{S}_n\mathbf{E}\mathbf{S}_n$), the computation essentially reduces to matrix-vector multiplication which is of complexity $O(g_n^2)$. Therefore, the complexity of computing $\{{\mathbf{S}_n^\prime}^k\}_{k=1}^K$ is $O(g_n^2)$ and the overall computational complexity of obtaining $\mathbf{z}_n^\prime$ is $O(d g_n^2)$.

Lastly, if $h(\lambda)$ is an arbitrary function, and we need to recompute the eigenvalue decomposition of $\mathbf{S}_n^\prime$, the problem is related to a classical problem termed ``downdating of the singular value decomposition'' of a perturbed matrix~\citep{gu1995downdating}. The overall complexity of obtaining $\mathbf{z}_n^\prime$ then becomes $O(g_n^2(\log^2 \xi + d))$, where $\xi$ is a parameter related to machine precision. 

It is worth pointing out that $O(g_n^2)$ is order-optimal with respect to the unlearning complexity of removing nodes from a graph $\mathcal{G}_n$, since the complexity of the basic operation, graph convolution (i.e., $\mathbf{S}\mathbf{x}$), is $O(g_n^2)$. As we will show in Section~\ref{sec:exp}, the unlearning complexity of using nontrainable GSTs is significantly smaller than that of using GNNs when constructing graph embeddings in the worst case. This is due to the fact that we may need to retrain GNNs frequently to eliminate the effect of removed nodes on the embedding procedure; on the other hand, we only need to recompute the embeddings of affected training graphs when using GSTs. The GSTs for different training graphs are computed independently, which may be seen as a form of sharding with small components. However, unlike traditional sharding-based methods~\citep{bourtoule2021machine,chen2021graph}, we do not need to carefully select the partition, and the sizes of the shards do not affect the performance of the final model.

\textbf{Using differentially-private GNNs for graph embeddings.} To ensure that the gradient residual norm does not grow excessively, we need to have control over the graph embedding procedure so that the embedding is stable with respect to small perturbations in the graph topology and features. The nontrainable GST is one choice, but DP-GNNs can also be used for generating the graph embeddings as they may improve the overall performance of the learner. Based on Theorem 5 from~\cite{guo2020certified}, the overall learning framework still satisfies the certified approximate removal criterion, and thus can be used as an approximate unlearning method as well. However, most DP-GNNs focus on node classification instead of graph classification tasks, and it remains an open problem to design efficient GNNs for graph classification problems while preserving node-level privacy. Moreover, it has been shown in~\cite{chien2023efficient} that DP-GNNs often require a high ``privacy cost'' ($\epsilon\geq 5$) (see Equation~(\ref{eq:CR_def})) to unlearn one node without introducing significant negative effects on model performance. In contrast, we find that in practice, our proposed unlearning approach based on GSTs only requires $\epsilon=1$. Therefore, using DP-GNNs for graph embedding in unlearning frameworks may not offer any advantages compared to alternatives.

\vspace{-0.05in}
\section{Experimental Results}\label{sec:exp}
\begin{figure*}[t]
    \centering
    \includegraphics[trim={0cm 0cm 0cm 0},clip,width=0.96\linewidth]{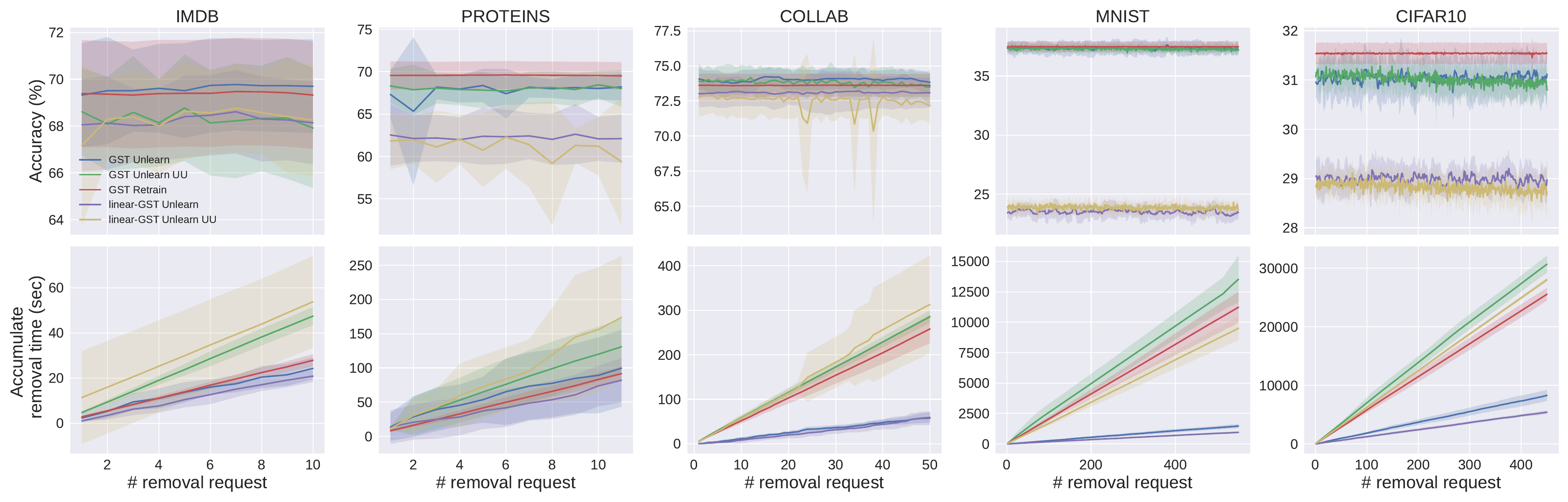}
    \vspace{-0.15in}
    \caption{Sequential unlearning results. We unlearn one node in each of the $10\%$ of the selected training graphs. The shaded area indicates one standard deviation. All approximate unlearning methods satisfy $(1,10^{-4})$-certified approximate removal.}
    \label{fig:Exp2}
\end{figure*}

\textbf{Settings.} We test our methods on five benchmarking datasets for graph classification, including two real social networks datasets IMDB, COLLAB~\citep{morris2020tudataset}, and three other standard graph classification benchmarking datasets MNIST, CIFAR10, PROTEINS~\citep{dobson2003distinguishing,krizhevsky2009learning,deng2012mnist,yanardag2015deep,dwivedi2020benchmarking}. As we focus on the limited training data regime, we use $10$ random splits for all experiments with the training/validation/testing ratio $0.1/0.1/0.8$. Following~\cite{guo2020certified}, we use LBFGS as the optimizer for all non-GNN methods due to its high efficiency on strongly convex problems. We adopt the Adam~\citep{kingma2014adam} optimizer for GNNs following the implementation of Pytorch Geometric library benchmarking examples~\citep{fey2019fast}. We compare our unlearning approach (Figure~\ref{fig:unlearn_objective} (a)) with a naive application of~\cite{guo2020certified} (Figure~\ref{fig:unlearn_objective} (b)) as well as complete retraining. The tested backbone graph learning models include GST, GFT, linear-GST (i.e., GST without nonlinear activations) and GIN~\citep{xu2018how}. For all approximate unlearning methods we use $(\epsilon,\delta) = (1,10^{-4})$ and noise $\alpha=0.1$ as the default parameters unless specified otherwise. The shaded area in all plots indicates one standard deviation. Additional details are in Appendix~\ref{app:Exp_detail}.

\textbf{Performance of the backbone models. }We first test the performance of all backbone graph learning models on the standard graph classification problem. The results are presented in Tables~\ref{tab:exp1-acc} and~\ref{tab:exp1-times}. We observe that GST has consistently smaller running times compared to GIN while offering matching or better accuracy. This validates the findings of~\cite{gao2019geometric} and confirms that GST is indeed efficient and effective in the limited training data regime. Compared to linear-GSTs, we find that the nonlinearity of GST is important to achieve better accuracy, with an average increase of $3.5\%$ in test accuracy over five datasets. In general, GST also significantly outperforms GFT with respect to accuracy with a significantly smaller running time. This is due to the fact that GST (with polynomial wavelet kernel functions as in~\cite{gao2019geometric}) does not require an eigenvalue decomposition as GFT does, which is computationally expensive to perform for large datasets.

\begin{table}[t]
\setlength{\tabcolsep}{5pt}
\centering
\caption{Test accuracy ($\%$) of the backbone graph learning methods for standard graph classification in the limited training data regime. The results report the mean accuracy and standard deviation. Bold numbers indicate the best results.}
\vspace{-0.1in}
\label{tab:exp1-acc}
\scriptsize
\begin{tabular}{@{}cccccc@{}}
\toprule
           & IMDB           & PROTEINS       & COLLAB         & MNIST                   & CIFAR10        \\ \midrule
GST & \textbf{68.56 $\pm$ 3.52} & \textbf{68.26 $\pm$ 2.28} & \textbf{74.42 $\pm$ 0.81} & 47.59 $\pm$ 0.25 & \textbf{33.12 $\pm$ 0.40} \\
linear-GST & \textbf{68.30 $\pm$ 3.67} & 62.79 $\pm$ 4.67 & 73.84 $\pm$ 0.70 & 38.52 $\pm$ 0.26          & 31.07 $\pm$ 0.21 \\
GFT        & 50.81 $\pm$ 1.32 & 49.67 $\pm$ 1.45 & 34.58 $\pm$ 0.79 & 10.13 $\pm$ 0.22          & 10.00 $\pm$ 0.15 \\
GIN        & 66.63 $\pm$ 4.29 & 65.12 $\pm$ 1.55 & 73.11 $\pm$ 1.43 & \textbf{48.17 $\pm$ 0.45} & 30.05 $\pm$ 0.59 \\ \bottomrule
\end{tabular}
\end{table}

\begin{table}[t]
\setlength{\tabcolsep}{4pt}
\centering
\caption{Running time (s) of the backbone graph learning methods for standard graph classification in the limited training data regime. The results report the mean accuracy and standard deviation. Bold numbers indicate the best results.}
\vspace{-0.1in}
\label{tab:exp1-times}
\scriptsize
\begin{tabular}{@{}cccccc@{}}
\toprule
    & IMDB           & PROTEINS       & COLLAB           & MNIST             & CIFAR10           \\ \midrule
GST        & 6.47 $\pm$ 0.89 & \textbf{7.57 $\pm$ 1.79} & \textbf{10.89 $\pm$ 1.08} & \textbf{82.94 $\pm$ 5.75} & \textbf{75.36 $\pm$ 1.17} \\
linear-GST & 6.92 $\pm$ 1.50 & \textbf{7.59 $\pm$ 1.25} & \textbf{10.94 $\pm$ 1.40} & \textbf{82.23 $\pm$ 0.83} & \textbf{74.98 $\pm$ 1.07} \\
GFT & \textbf{4.43 $\pm$ 1.04} & 9.00 $\pm$ 0.96  & 137.69 $\pm$ 1.29  & 1307.43 $\pm$ 1.10  & 4240.62 $\pm$ 2.56  \\
GIN & 23.13 $\pm$ 1.32         & 21.94 $\pm$ 0.92 & 949.06 $\pm$ 63.68 & 1279.26 $\pm$ 30.92 & 1239.03 $\pm$ 33.06 \\ \bottomrule
\end{tabular}
\end{table}

\begin{figure}[h]
    \centering
    \includegraphics[trim={0cm 0cm 0cm 0},clip,width=0.95\linewidth]{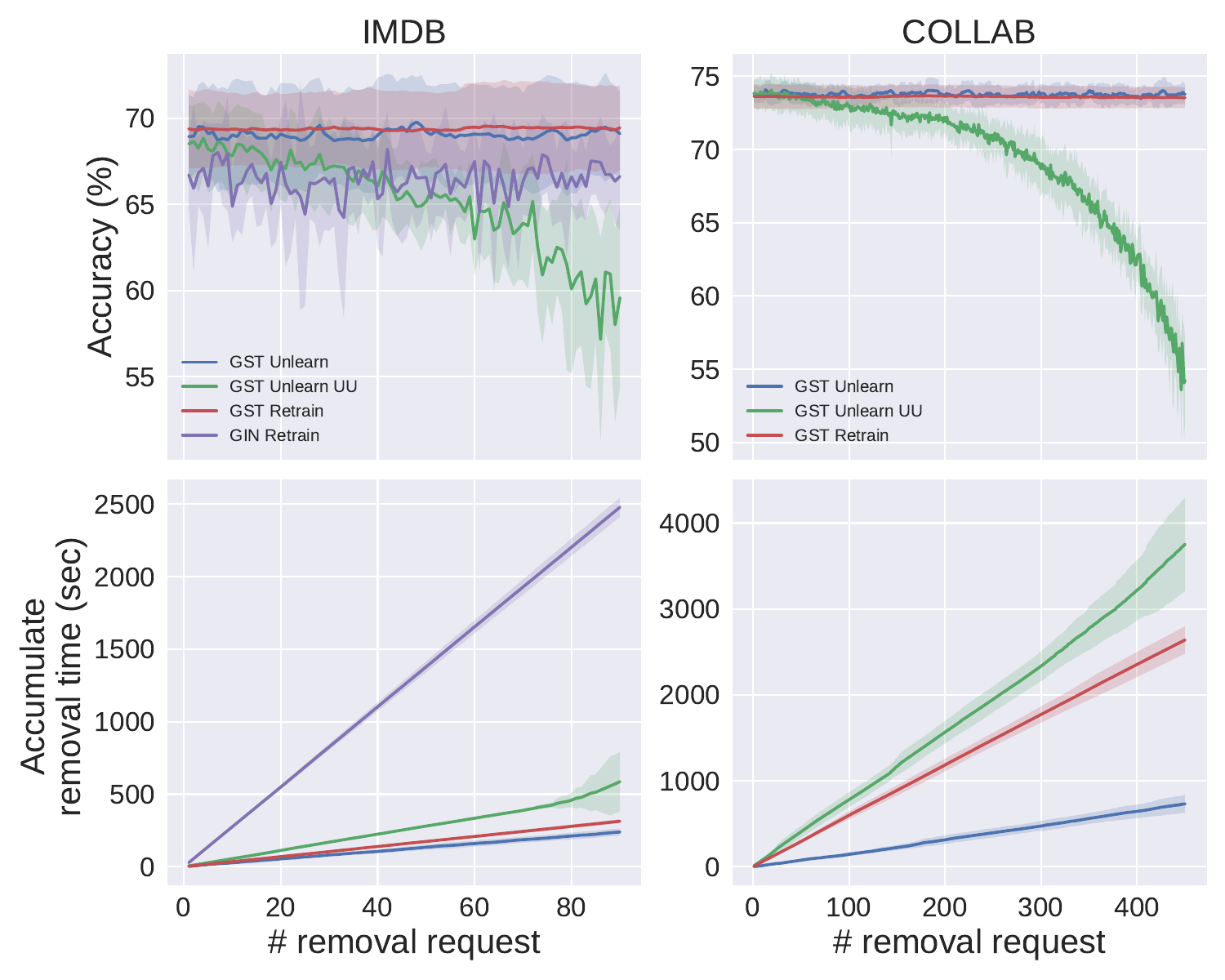}
    \vspace{-0.15in}
    \caption{Unlearning results for the extreme setting. We unlearn one node in each of the $90\%$ training graphs.}
    \label{fig:Exp3}
\end{figure}

\textbf{Performance of different unlearning methods. }Next, we test various unlearning schemes combined with GST. In this set of experiments, we sequentially unlearn one node from each of the selected $10\%$ training graphs. We compare our Algorithm~\ref{alg:unlearning} with the unstructured unlearning method~\citep{guo2020certified} and complete retraining (Retrain). Note that in order to apply unstructured unlearning to the graph classification problem, we have to remove the entire training graph whenever we want to unlearn even one single node from it. The results are depicted in Figure~\ref{fig:Exp2}. We can see that our unlearning scheme with GSTs has accuracy comparable to that of complete retraining but much lower time complexity. Also, note that a naive application of~\cite{guo2020certified} (indexed ``UU'' for unstructured unlearning) results in complete retraining in almost all the cases (see Table~\ref{tab:num_retrain_short}). In addition, the method requires removing the entire training graph instead of just one node as requested, thus the accuracy can drop significantly when unlearning many requests (Figure~\ref{fig:Exp3}).

\begin{table}[h]
\setlength{\tabcolsep}{4pt}
\centering
\caption{Number of retraining from scratch during sequential unlearning one node from $10\%$ of training graphs.}
\vspace{-0.1in}
\label{tab:num_retrain_short}
\small
\begin{tabular}{@{}cccccc@{}}
\toprule
               & IMDB & PROTEINS & COLLAB & MNIST & CIFAR10 \\ \midrule
GST            & 3.3           & 7.2               & 7.7             & 65.9           & 113.0            \\
GST UU        & 10.0          & 11.0              & 50.0            & 550.0          & 450.0            \\
GST Retrain    & 10            & 11                & 50              & 550            & 450              \\
linear-GST     & 3.0           & 6.8               & 6.3             & 54.1           & 91.6             \\
linear-GST UU & 10.0          & 11.0              & 49.6            & 532.5          & 450.0            \\ \bottomrule
\end{tabular}
\end{table}

We further examine the performance of these unlearning approaches in the ``extreme'' graph unlearning setting: Now we unlearn one node from each of the $90\%$ training graphs sequentially. Due to the high time complexity of baseline methods, we conduct this experiment only on one small dataset, IMDB, and one medium dataset, COLLAB. We also compare completely retraining GIN on IMDB. The results are shown in Figure~\ref{fig:Exp3}. We observe that retraining GIN is indeed prohibitively complex in practice. A naive application of~\cite{guo2020certified} (indexed ``UU'') leads to a huge degradation in accuracy despite the high running complexity. This is due to the fact that one has to remove the entire training graph for each node unlearning request, which is obviously wasteful. Overall, our proposed strategy combined with GST gives the best results regarding time complexity and offers comparable test accuracy and affordable privacy costs.

\begin{figure}[h]
    \centering
    \includegraphics[trim={0cm 0cm 0cm 0},clip,width=0.95\linewidth]{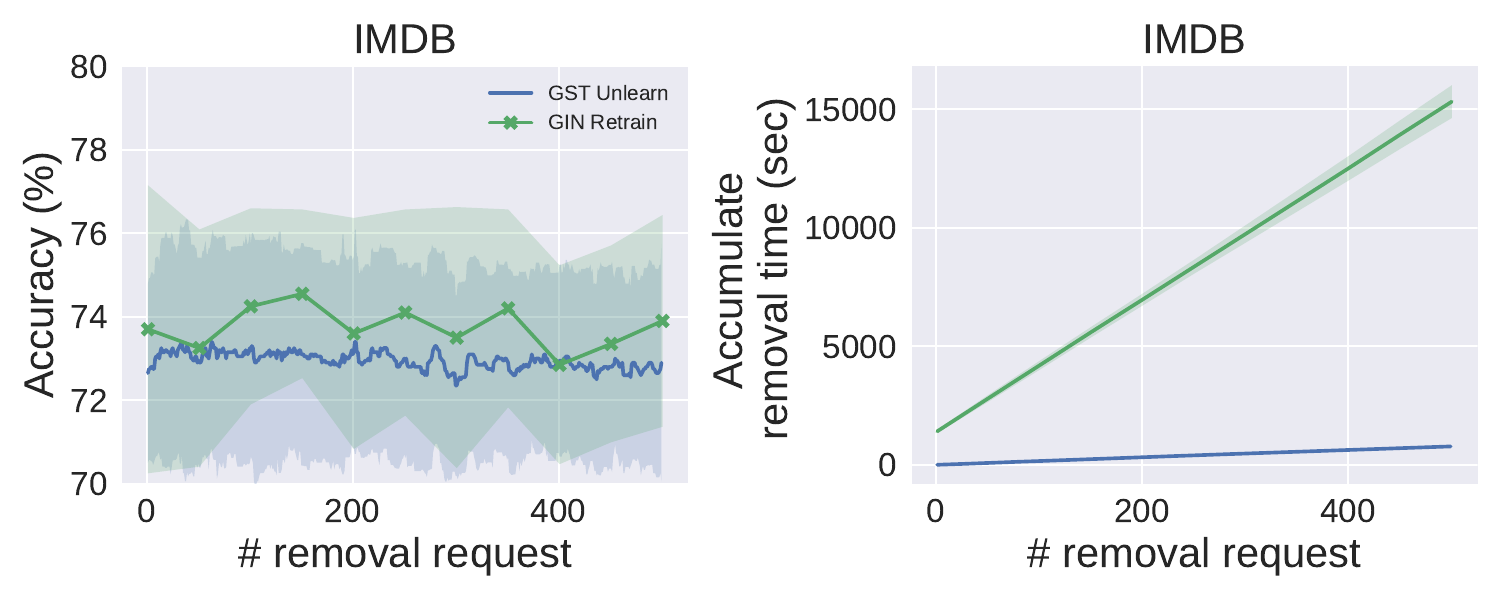}
    \vspace{-0.15in}
    \caption{Unlearning results when training data is abundant. We unlearn one node in each of the $83\%$ training graphs.}
    \label{fig:Exp4}
\end{figure}

\vspace{-0.1in}
\textbf{Performance of the proposed method with abundant training data.} Next, we use the IMDB dataset as an example to demonstrate the performance of our unlearning method compared to complete retraining; the training/validation/testing ratio is set to $0.6/0.2/0.2$, and out of $600$ training graphs, we unlearn $500$ samples in terms of each removing a single node. The results (see Figure~\ref{fig:Exp4}) show that our approach still offers roughly a $10$-fold decrease in unlearning time complexity compared to retraining GIN, with comparable test accuracy. Note that due to the high complexity of retraining GIN in this setting, we only performed complete retraining for the number of removal requests indicated using green marks in Figure~\ref{fig:Exp4}. The green lines correspond to the interpolated results.

\begin{figure}[h]
    \centering
    \includegraphics[trim={0cm 0cm 0cm 0},clip,width=0.95\linewidth]{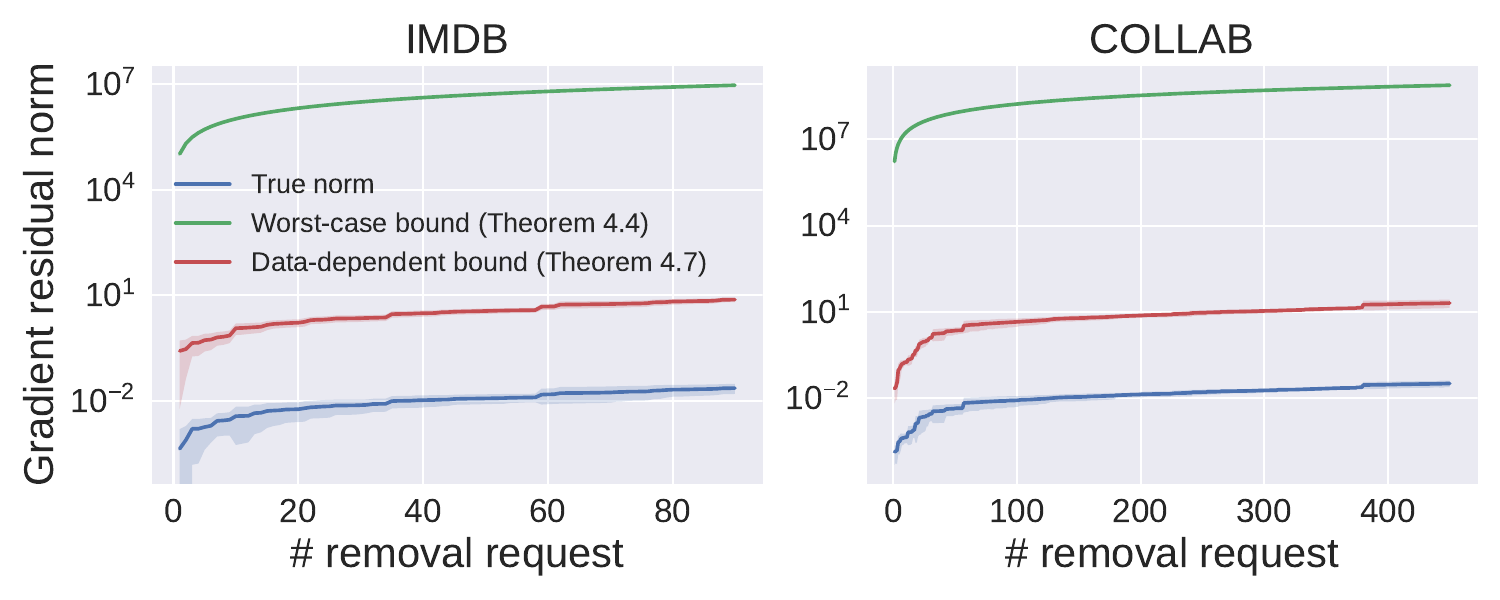}
    \vspace{-0.15in}
    \caption{Comparison of the worst-case and data-dependent bounds on the gradient residual norm, and the true value of the gradient residual norm.}
    \label{fig:gnorm}
\end{figure}

\textbf{Bounds on the gradient residual norm.} We also examine the \emph{worst-case} bounds (Theorem~\ref{thm:worst_case_single}) and the \emph{data-dependent} bounds (Theorem~\ref{thm:data_dependent}) of Algorithm~\ref{alg:unlearning} computed during the unlearning process, along with the true value of the gradient residual norm (True norm) to validate our theoretical findings from Section~\ref{sec:gc_unlearn}. For simplicity, we set $\alpha = 0$ during training. Figure~\ref{fig:gnorm} confirms that the worst-case bounds are looser than the data-dependent bounds.

\vspace{-0.05in}
\section{Conclusion}\label{sec:conclusion}
We studied the first nonlinear graph learning framework based on GSTs that accommodates an approximate unlearning mechanism with provable performance guarantees on computational complexity. With the adoption of the mathematically designed transform GSTs, we successfully extended the theoretical analysis on linear models to nonlinear graph learning models. Our experimental results validated the remarkable unlearning efficiency improvements compared to complete retraining.

\vspace{-0.05in}
\begin{acks}
This work was funded by NSF grants 1816913 and 1956384.
\end{acks}

\clearpage
\bibliographystyle{ACM-Reference-Format}
\bibliography{sample-base}

\clearpage
\appendix

\section{Proof of Theorem~\ref{thm:worst_case_single_feature}}\label{app:pf_single_feature_worst}
Our proof is a generalization of the proof in~\cite{guo2020certified}. Due to space limitations, we show proof sketches here and delegate the details to a full version of this paper. Let $G(\mathbf{w}) = \nabla L(\mathbf{w},\mathcal{D}^\prime)$ denote the gradient of the empirical risk on the updated dataset $\mathcal{D}^\prime$ at $\mathbf{w}$. By Taylor's expansion theorem, 
there exists some $\eta\in [0,1]$ such that
\begin{align}
    G(\mathbf{w}^{\prime}) &= G(\mathbf{w}^\star + \mathbf{H}_{\mathbf{w}^\star}^{-1}\Delta) = G(\mathbf{w}^\star) + \nabla G(\mathbf{w}^\star+\eta\mathbf{H}_{\mathbf{w}^\star}^{-1}\Delta)\mathbf{H}_{\mathbf{w}^\star}^{-1}\Delta \notag \\
    & = (\mathbf{H}_{\mathbf{w}_\eta}-\mathbf{H}_{\mathbf{w}^\star})\mathbf{H}_{\mathbf{w}^\star}^{-1}\Delta,
\end{align}
where $\mathbf{H}_{\mathbf{w}_\eta} \triangleq \nabla G(\mathbf{w}^\star+\eta\mathbf{H}_{\mathbf{w}^\star}^{-1}\Delta)$, which is the Hessian at $\mathbf{w}_\eta \triangleq \mathbf{w}^\star+\eta\mathbf{H}_{\mathbf{w}^\star}^{-1}\Delta$. By the Cauchy-Schwartz inequality, we have
\begin{align}\label{eq:gr_norm}
    \|G(\mathbf{w}^\prime)\| \leq \|\mathbf{H}_{\mathbf{w}_\eta}-\mathbf{H}_{\mathbf{w}^\star}\| \|\mathbf{H}_{\mathbf{w}^\star}^{-1}\Delta\|.
\end{align}
Note that since $\ell^{\prime\prime}$ is $\gamma_2$-Lipschitz, for $\forall i\in[n]$ we have 
\begin{align}\label{eq:hessian_bound_single}
\left\|\nabla^2 \ell\left(\mathbf{w}_\eta^{T} \mathbf{z}_i, y_i\right)-\nabla^2 \ell\left(\left(\mathbf{w}^\star\right)^{T} \mathbf{z}_i, y_i\right)\right\|
\leq \gamma_2 F^3 \left\|\mathbf{H}_{\mathbf{w}^\star}^{-1} \Delta\right\|,
\end{align}
where $F=\sqrt{\sum_{l=0}^{L-1} B^{2l}}$. The upper bound comes from the frame property of graph wavelets. More specifically, the sum of energy for the $l$-th layer in the scattering tree is upper bounded by $B^{2l}\|\mathbf{x}_i\|^2$, where $\mathbf{x}_i$ is the input 
graph signal at the $0$-th layer and $\|\mathbf{x}_i\|^2$ is the corresponding energy. Suppose that we are using the low-pass averaging operator for $U_i$ in GST (i.e., $U_i=\frac{1}{g_i}\mathbf{1}^T, \|U_i\|=\frac{1}{\sqrt{g_i}}$). Then 
\begin{align}
    \left\|\mathbf{z}_i\right\|^2 =\sum_{p_j(l)} \phi_{p_j(l)}^2(\mathbf{S}_i, \mathbf{x}_i) \leq \frac{\sum_{l=0}^{L-1} B^{2l}\|\mathbf{x}_i\|^2}{g_i}\leq \sum_{l=0}^{L-1} B^{2l},
\end{align}
since $\|\mathbf{x}_i\|^2 \leq g_i$ based on Assumption~\ref{asp:guo}. Therefore, we have $\left\|\mathbf{z}_i\right\|^3\leq \left(\sum_{l=0}^{L-1} B^{2l}\right)^{3/2}=F^3$. Summing the expression in Equation~(\ref{eq:hessian_bound_single}) over the updated dataset $\mathcal{D}^\prime$, we conclude that
$$
\|\mathbf{H}_{\mathbf{w}_\eta}-\mathbf{H}_{\mathbf{w}^\star}\| \leq \gamma_2 n F^3 \left\|\mathbf{H}_{\mathbf{w}^\star}^{-1} \Delta\right\|.
$$
Replacing the above equality into Equation~(\ref{eq:gr_norm}), we arrive at 
$$
\|G(\mathbf{w}^\prime)\|\leq \gamma_2 n F^3 \left\|\mathbf{H}_{\mathbf{w}^\star}^{-1} \Delta\right\|^2.
$$

Next, we bound $\|\mathbf{H}_{\mathbf{w}^\star}^{-1}\Delta\|$. Since $L(\cdot,\mathcal{D}^\prime)$ is $(\lambda n)$-strongly convex, we have $\|\mathbf{H}_{\mathbf{w}^\star}^{-1}\| \leq \frac{1}{\lambda n}$. For $\Delta$, by definition,
\begin{align*}
\Delta=\nabla\ell\left(\left(\mathbf{w}^\star\right)^T\mathbf{z}_n,y_n\right) - \nabla\ell\left(\left(\mathbf{w}^\star\right)^T\mathbf{z}_n^\prime,y_n\right).
\end{align*}
There are two ways to bound $\|\Delta\|$. First, based on Assumption~\ref{asp:guo},
$$
    \|\Delta\|\leq \left\|\nabla\ell\left(\left(\mathbf{w}^\star\right)^T\mathbf{z}_n,y_n\right)\right\| + \left\|\nabla\ell\left(\left(\mathbf{w}^\star\right)^T\mathbf{z}_n^\prime,y_n\right)\right\| \leq 2C_1.
$$
Second, we can also bound $\|\Delta\|$ by 
\begin{align*}
    \|\Delta\|\leq &\left|\ell^\prime\left(\left(\mathbf{w}^\star\right)^T\mathbf{z}_n,y_n\right) - \ell^\prime\left(\left(\mathbf{w}^\star\right)^T\mathbf{z}_n^\prime,y_n\right)\right| \|\mathbf{z}_n\| \\
    &+ \left|\ell^\prime\left(\left(\mathbf{w}^\star\right)^T\mathbf{z}_n^\prime,y_n\right)\right|\|\mathbf{z}_n-\mathbf{z}_n^\prime\| \\
    \leq &\gamma_1\|\mathbf{z}_n-\mathbf{z}_n^\prime\|\|\mathbf{w}^\star\|\|\mathbf{z}_n\| + C_2\|\mathbf{z}_n-\mathbf{z}_n^\prime\|.
\end{align*}
For $\|\mathbf{w}^\star\|$, we have
$$
    0=\nabla L(\mathbf{w}^\star,\mathcal{D})=\sum_{i=1}^n \nabla\ell \left(\left(\mathbf{w}^\star\right)^T\mathbf{z}_i,y_i\right) + \lambda n\mathbf{w}^\star,
$$
which leads to
$$
    \|\mathbf{w}^\star\|\leq \frac{\sum_{i=1}^n\|\nabla\ell \left(\left(\mathbf{w}^\star\right)^T\mathbf{z}_i,y_i\right)\|}{\lambda n}\leq \frac{C_1}{\lambda}.
$$
From Theorem 1 of~\cite{pan2021spatiotemporal}, by setting $T=1$, we have
$$
    \|\mathbf{z}_n-\mathbf{z}_n^\prime\| \leq \sqrt{\frac{\sum_{l=0}^{L-1} B^{2l}}{g_n}}\|\mathbf{x}_n-\mathbf{x}_n^\prime\|\leq \frac{F}{\sqrt{g_n}}.
$$
Combining the above results we obtain
$$
\|\Delta\| \leq \frac{\gamma_1 C_1 F^2 + \lambda C_2 F}{\lambda\sqrt{g_n}}.
$$
Therefore, 
\begin{align}\label{eq:feature_choose_min}
    \|\mathbf{H}_{\mathbf{w}^\star}^{-1}\Delta\|\leq \frac{1}{\lambda n}\cdot \min\left\{2C_1,\frac{\gamma_1 C_1 F^2 + \lambda C_2 F}{\lambda\sqrt{g_n}}\right\}.
\end{align}
Replacing Equation~(\ref{eq:feature_choose_min}) into the bound on $\|G(\mathbf{w}^\prime)\|$, we arrive at
$$
\|G(\mathbf{w}^\prime)\| \leq \frac{\gamma_2 F^3}{\lambda^2 n}\min\left\{4C_1^2,\frac{(\gamma_1 C_1 F^2 + \lambda C_2 F)^2}{\lambda^2 g_n}\right\},
$$
which completes the proof.

\section{Proof of Theorem~\ref{thm:worst_case_single}}\label{app:pf_single_worst}
Following the same proof approach as described in Appendix~\ref{app:pf_single_feature_worst}, for the case of single node removal we have 
$$
\|G(\mathbf{w}^\prime)\|\leq \gamma_2 n F^3 \left\|\mathbf{H}_{\mathbf{w}^\star}^{-1} \Delta\right\|^2.
$$
The main difference between feature removal and node removal proof is that the norm of the change in the graph embeddings $\|\mathbf{z}_n^\prime-\mathbf{z}_n\|$ obtained by GST with respect to structural perturbation is proportional to the norm of the entire graph signal $\|\mathbf{x}_n\|$. More specifically, when the magnitude of the structural perturbation is controlled and the 
wavelet kernel functions satisfy certain mild conditions, from Theorem 2 of~\cite{pan2021spatiotemporal} we have that (for $T=1$)
\begin{align*}
\|\mathbf{z}_n^\prime-\mathbf{z}_n\| &= \left\|\Phi(\mathbf{S}_n^\prime, \mathbf{x}_n^\prime)-\Phi(\mathbf{S}_n, \mathbf{x}_n)\right\| \\
&= \left\|\Phi(\mathbf{S}_n^\prime, \mathbf{x}_n^\prime) - \Phi(\mathbf{S}_n^\prime, \mathbf{x}_n) + \Phi(\mathbf{S}_n^\prime, \mathbf{x}_n) - \Phi(\mathbf{S}_n, \mathbf{x}_n)\right\| \\
&\leq \frac{F}{\sqrt{g_n}} + O\left(\sqrt{\frac{\sum_{l=0}^{L-1} l^2(B^2J)^l}{B^2 g_n}}\right)\|\mathbf{x}_n\| \\
&\leq \frac{F}{\sqrt{g_n}} + O\left(\sqrt{\frac{\sum_{l=0}^{L-1} l^2(B^2J)^l}{B^2}}\right),
\end{align*}
where $\|\mathbf{x}_n\|\leq \sqrt{g_n}$ and $F=\sqrt{\sum_{l=0}^{L-1} B^{2l}}$. In this case, the second term in the upper bound of $\|\mathbf{z}_n^\prime-\mathbf{z}_n\|$ does not decrease when $g_n$ increases, and $\|\Delta\|\leq 2C_1$ is very likely a tighter upper bound than the other option. Therefore, we have $\|G(\mathbf{w}^\prime)\| \leq \frac{4\gamma_2 C_1^2 F^3}{\lambda^2 n}$, which completes the proof.

\section{Proof of Corollaries~\ref{coro:feature_worst_case_batch} and~\ref{coro:worst_case_batch}}\label{app:pf_coro_worst}
The proof of Corollaries~\ref{coro:feature_worst_case_batch} and~\ref{coro:worst_case_batch} follows along similar lines as that in Appendix~\ref{app:pf_single_feature_worst} and~\ref{app:pf_single_worst}. With the same argument, we can show that $\|G(\mathbf{w}^\prime)\|\leq \gamma_2 n F^3 \left\|\mathbf{H}_{\mathbf{w}^\star}^{-1} \Delta\right\|^2$. The only difference arises from the fact that there could be now at most $2m$ terms in $\Delta$ instead of just $2$ terms, and the worst case arises when each of these $m$ nodes that requested removal comes from a different graph. In this case, $\|\Delta\|\leq 2mC_1$. Therefore, for batch feature removal we have
$$
\|\nabla L(\mathbf{w}^\prime,\mathcal{D}^\prime)\|\leq \frac{\gamma_2 m^2 F^3}{\lambda^2 n}\min\left\{4C_1^2,\frac{(\gamma_1 C_1 F^2 + \lambda C_2 F)^2}{\lambda^2 g_n}\right\},
$$
while for batch node removal, we have
$$
\|\nabla L(\mathbf{w}^\prime,\mathcal{D}^\prime)\| \leq \frac{4\gamma_2 m^2 C_1^2 F^3}{\lambda^2 n}.
$$
Note that we require $m<\min_i g_i$ not to unlearn the entire graph, otherwise the number of training samples in Equation~(\ref{eq:loss}) would be less than $n$ and tighter bounds on gradient residual norm can be derived. 
Nevertheless, if we indeed need to unlearn multiple graphs completely, we can always unlearn them first based on the batch removal procedure in~\cite{guo2020certified}, and then perform our unlearning procedure based on Equation~(\ref{eq:update_rule}) for the remaining graphs.

\section{Proof of Theorem~\ref{thm:data_dependent}}\label{app:pf_data_dependent}
Based on the loss function defined in Equation~(\ref{eq:loss}), the Hessian of $L(\cdot,\mathcal{D}^\prime)$ at $\mathbf{w}$ takes the form $\mathbf{H}_{\mathbf{w}}=\left(\mathbf{Z}^\prime\right)^T\mathbf{D}_{\mathbf{w}}\mathbf{Z}^\prime + \lambda n \mathbf{I}_d$, where $\mathbf{Z}^\prime\in\mathbb{R}^{n\times d}$ is the data matrix corresponding to $\mathcal{D}^\prime$ and $\mathbf{D}_{\mathbf{w}}$ denotes the diagonal matrix with diagonal values $(\mathbf{D}_{\mathbf{w}})_{ii}=\ell^{\prime\prime}(\mathbf{w}^T\mathbf{z}_i,y_i)$.

From the proof of Theorem~\ref{thm:worst_case_single} we know that 
\begin{align}\label{eq:data_dependent_intermediate}
\left\|\nabla L\left(\mathbf{w}^{\prime}, \mathcal{D}^{\prime}\right)\right\| & \leq \left\|\left(\mathbf{H}_{\mathbf{w}_\eta}-\mathbf{H}_{\mathbf{w}^\star}\right) \mathbf{H}_{\mathbf{w}^\star}^{-1}\Delta\right\| \notag\\
&= \left\|\left(\mathbf{Z}^\prime\right)^T(\mathbf{D}_{\mathbf{w}_\eta} - \mathbf{D}_{\mathbf{w}^\star})\mathbf{Z}^\prime \mathbf{H}_{\mathbf{w}^\star}^{-1}\Delta\right\|\notag\\
& \leq \|\mathbf{Z}^\prime\| \|\mathbf{D}_{\mathbf{w}_\eta} - \mathbf{D}_{\mathbf{w}^\star}\| \|\mathbf{Z}^\prime \mathbf{H}_{\mathbf{w}^\star}^{-1}\Delta\|,
\end{align}
where $\mathbf{w}_\eta \triangleq \mathbf{w}^\star+\eta\mathbf{H}_{\mathbf{w}^\star}^{-1}\Delta$ for some $\eta\in[0,1]$. Since $\mathbf{D}_{\mathbf{w}_\eta} - \mathbf{D}_{\mathbf{w}^\star}$ is a diagonal matrix, its operator norm corresponds to the maximum absolute value of the diagonal elements. In the proof of Theorem~\ref{thm:worst_case_single} we showed that for $\forall i\in[n]$, 
\begin{align*}
\left|\ell^{\prime \prime}\left(\mathbf{w}_\eta^{T} \mathbf{z}_i, y_i\right)-\ell^{\prime \prime}\left(\left(\mathbf{w}^\star\right)^{T} \mathbf{z}_i, y_i\right)\right| &\leq \gamma_2 \|\mathbf{w}_\eta - \mathbf{w}^\star\| \|\mathbf{z}_i\| \\
&\leq \gamma_2 F \left\|\mathbf{H}_{\mathbf{w}^\star}^{-1} \Delta\right\|.
\end{align*}
Thus we have that $\|\mathbf{D}_{\mathbf{w}_\eta} - \mathbf{D}_{\mathbf{w}^\star}\|\leq \gamma_2 F \left\|\mathbf{H}_{\mathbf{w}^\star}^{-1} \Delta\right\|$. Combining this result with Equation~(\ref{eq:data_dependent_intermediate}) completes the proof. Note that this analysis holds for both single-removal and batch-removal, as well as both feature and node removal requests, since it does not require an explicit upper bound on the norm of gradient change $\|\Delta\|$.

\section{Choices of Graph Wavelets}\label{app:wavelet_formulation}
There are many off-the-shelf graph wavelets we can choose from. They are mainly used for extracting features from multiple frequency bands of the input signal spectrum. Some are listed below.

\textbf{Monic Cubic wavelets.} Monic Cubic wavelets~\citep{hammond2011wavelets} use a kernel function $h(\lambda)$ of the form
$$
h(\lambda) = 
    \begin{cases}
    \lambda& \text{for}\quad \lambda < 1;\\
    -5+11\lambda-6\lambda^2+\lambda^3& \text{for}\quad 1 \leq\lambda \leq 2; \\
    2/\lambda &\text{for}\quad \lambda > 2.
    \end{cases}
$$
Different scalings of the filters are implemented by scaling and translating the above kernel function.

\textbf{Itersine wavelets.} Itersine wavelets define the kernel function at scale $j$ as 
$$
h_j(\lambda)=\sin\left(\frac{\pi}{2}\cos^2\left(\pi\left(\lambda-\frac{j-1}{2}\right)\right)\right)\cdot \mathbb{I}\left[\frac{j}{2}-1\leq\lambda\leq \frac{j}{2}\right].
$$
Itersine wavelets form tight and energy-preserving frames.

\textbf{Diffusion scattering wavelets.} A diffusion scattering wavelet filter bank~\citep{gama2018diffusion} contains a set of filters based on a lazy diffusion matrix $\mathbf{S}=\frac{1}{2}(\mathbf{I}+\mathbf{D}^{-1/2}\mathbf{A}\mathbf{D}^{-1/2})$, where $\mathbf{A}$ is the adjacency matrix and $\mathbf{D}$ is the corresponding degree matrix. The filters are defined as
\begin{align*}
    \mathbf{H}_j(\mathbf{S}) = \begin{cases}
    \mathbf{I} - \mathbf{S},\; &j = 0,\\
    \mathbf{S}^{{2}^{j-1}} - \mathbf{S}^{{2}^{j}},\; &j > 0.
    \end{cases}
\end{align*}
Note that for diffusion scattering the low-pass operator $U$ is defined as $\mathbf{d}/\|\mathbf{d}\|_1$, where $\mathbf{d}$ is the diagonal of $\mathbf{D}$.

\textbf{Geometric scattering wavelets.} The definition of geometric scattering~\citep{gao2019geometric} is similar as diffusion scattering, except that the lazy random walk matrix used in geometric scattering is defined as $\mathbf{S}=\frac{1}{2}(\mathbf{I}+\mathbf{A}\mathbf{D}^{-1})$. And geometric scattering will also record different moments of filtered signals $\mathbf{H}_j^q(\mathbf{x}), \forall q\in[Q]$ as features.

Note that one is also allowed to customize the graph wavelets, as long as they satisfy the constraint
\begin{align*}
    A^{2}\|\mathbf{x}\|^{2} \leq \sum_{j=1}^{J}\left\|\mathbf{H}_{j}\mathbf{x}\right\|^{2} \leq B^{2}\|\mathbf{x}\|^{2},\quad |\lambda h'(\lambda)|\leq \text{const}~~ \forall \lambda,
\end{align*}
where $A, B$ are scalar constants and $h(\cdot)$ is the kernel function.

\section{Additional Experimental Details}\label{app:Exp_detail}
\textbf{Hyperparameters.} We follow the PyG benchmarking code to preprocess the datasets. For datasets without node
features, we generate synthetic node features based on node degrees. For all methods, we perform training with $500$ epochs. For GIN, we tune the 
hyperparameters on the small dataset IMDB and subsequently use them on all other datasets. We use $2$ layers, $64$ hidden dimensions and a learning rate $10^{-4}$ in the Adam optimizer. We find that this setting works well in general. For GST, we use geometric scattering wavelets in the graph embedding procedure and fix the learning rate of the LBFGS optimizer to $0.5$ for training the classifier. Other hyperparameters used for GST are described in Table~\ref{tab:hyperparam}. Here $J$ represents the number of scales for graph wavelets $\{h_j\}_{j=1}^J$, $L$ represents the number of layers in the scattering tree (with the root node at layer $0$), and $Q$ represents the number of moments computed for geometric scattering wavelets (see Appendix~\ref{app:wavelet_formulation}).


\begin{table}[h]
\centering
\caption{The hyperparameters of GST for all experiments.}
\label{tab:hyperparam}
\begin{tabular}{@{}ccccc|ccccc@{}}
\toprule
            & \multicolumn{4}{c|}{\begin{tabular}[x]{@{}c@{}}Standard graph\\ classification \end{tabular}} 
            & \multicolumn{5}{c}{\begin{tabular}[x]{@{}c@{}}Unlearning graph\\ classification \end{tabular}} \\
            & $J$        & $Q$        & $L$        & $\lambda$        & $J$     & $Q$     & $L$     & $\lambda$    & $\alpha$     \\\midrule
IMDB & 5        & 4        & 3        & $10^{-4}$        & 4     & 3     & 3     & $10^{-3}$    & $10^{-1}$    \\
PROTEINS    & 5        & 4        & 3        & $10^{-4}$        & 5     & 4     & 3     & $10^{-4}$    & $10^{-1}$    \\
COLLAB      & 3        & 3        & 2        & $10^{-4}$        & 3     & 3     & 2     & $10^{-4}$    & $10^{-1}$    \\
MNIST       & 5        & 4        & 3        & 0                & 5     & 4     & 3     & $10^{-6}$    & $10^{-1}$    \\
CIFAR10     & 5        & 4        & 3        & 0                & 5     & 4     & 3     & $10^{-6}$    & $10^{-1}$    \\ \bottomrule
\end{tabular}
\end{table}

\end{document}